# Reliability Assessment of Distribution System Using Fuzzy Logic for Modelling of Transformer and Line Uncertainties

Ahmad Shokrollahi[1], Hossein Sangrody[2], *Student Member, IEEE,* Mahdi Motalleb[3], *Student Member, IEEE*, Mandana Rezaeiahari[4], Elham Foruzan[5], *Student Member, IEEE*, Fattah Hassanzadeh[6]

[1]Mazandaran Regional Electric Company, Sari, Iran, Ashokrollahi@mazrec.co.ir
[2]Electrical and Computer Engineering Department, State University of New York at Binghamton, NY, USA,
[3]Mechanical Engineering, University of Hawaii at Manoa, HI, USA
[4]Industrial Engineering Department, State University of New York at Binghamton, NY, USA
[5]Electrical and Computer Engineering Department, University of Nebraska, NE, USA
[6]Electrical and Instrument Engineering Department, Mapna Turbine Engineering and Manufacturing Company, Karaj, Iran

*Abstract*—Reliability assessment of distribution system, based on historical data and probabilistic methods, leads to an unreliable estimation of reliability indices since the data for the distribution components are usually inaccurate or unavailable. Fuzzy logic is an efficient method to deal with the uncertainty in reliability inputs. In this paper, the ENS index along with other commonly used indices in reliability assessment are evaluated for the distribution system using fuzzy logic. Accordingly, the influential variables on the failure rate and outage duration time of the distribution components, which are natural or human-made, are explained using proposed fuzzy membership functions. The reliability indices are calculated and compared for different cases of the system operations by simulation on the IEEE RBTS Bus 2. The results of simulation show how utilities can significantly improve the reliability of their distribution system by considering the risk of the influential variables.

## I. Introduction

Energy supply without interruption is one the most major expectations of the customers of a power system and a lot of studies in planning, operating, and controlling fields endeavor directly or indirectly to meet such an expectation [1-4]. Practically, consistent supply is not possible. The reasons for such an issue are emanated from many natural or human-made causes. Adverse weather condition, flood, trees connection to the power network, improper maintenance of electric component, and improper management of power system are some examples of such causes.

Although the failure in generation and transmission can cause serious damage to the system, the failure rate in the distribution network is higher since it is physically more extensive than two other system levels and it has more components with lower protection and maintenance. In addition, the reliability of distribution system is vital to have better management on the distributed energy resources (DERs) and their intermittency [5]. Generally, the reliability of a power system is calculated according to failure rate and outage duration indices of its components. Such indices are usually calculated based on historical data and probabilistic method [6-8]. Some of the methods to derive the reliability indices based on historical data of component and probabilistic statistics are network reduction, frequency and duration, Markov modeling, and Monte Carlo-based method [9-13]. However, such methods depend on adequate and accurate archived database for each component in which a complete specification of each component explaining environmental and operational conditions, failure rate and reasons, component age, etc. are taken into consideration. Unfortunately, distribution system lacks such a comprehensive database, so reliability analysis based on the indices derived from inadequate or uncertain data results in inaccurate estimation. In addition, the failure rate is usually considered a constant value in the aforementioned methods whereas in practice the failure rate changes over time or in different environments [14]. Fuzzy logic is an efficient tool to deal with uncertainties. In [15-16] ], authors successfully utilized fuzzy methods to handle uncertainties and significantly improve their results.

Fuzzy sets theory, whose rules are defined based on human logic and experts' skill, can model the uncertainty, mathematically. In [17], a fuzzy logic based method is applied to calculate reliability indices for generation and transmission systems. In [18], fuzzy sets theory is applied in reliability assessment for planning purpose, and the highest risk feeders are determined for remedial actions. Although in [18] several influential factors are modeled using fuzzy logic, only influential variables on the feeder line are represented by fuzzy logic. In addition, the effect of weather condition which is one of the most significant reasons for the failure of feeder line is not considered in the fuzzy modeling. In this paper, fuzzy sets theory is applied to model the uncertainty of failure rate of the components in a distribution system in different environmental and operational conditions. The components considered in this study are line and transformer which are highly susceptible components in distribution networks. Using fuzzy logic, the mathematical representation of influential variables on the failure of the aforementioned components are derived and the simulation is done on the IEEE Reliability Test System (RBTS Bus 2) for different cases of influential variables [19]. The result of simulations shows the efficiency of the proposed fuzzy logic modeling in deriving reliability indices and decreasing the operational risk.

The rest of paper is organized as follows. In section II, the commonly used reliability indices and the fuzzy set theory is elaborated. Sections III represents the fuzzy models of influential variables on the reliability of the distribution network. The simulation results and the conclusion are represented in Section IV and V, respectively.

## II. RELIABILITY INDICES AND FUZZY SET

The distribution system is usually operated radially and it includes feeders, sectionalizing device, transformers, overhead lines or cables, loads, etc. [20]. A failure in the components of radial distribution system from feeder to the load point results in failure in supplying all or some customers of the feeders. In reliability studies, reliability indices of a feeder can be derived when two parameters of failure rate ($\lambda_i$) and average outage duration ($r_i$) are specified for each series component ($i$) from source to load point. Accordingly, in a load point, average failure rate ($\lambda$) which is the probability of failure, and average outage duration ($U$) are calculated as follow.

$$\lambda = \sum_{i=1}^{N} \lambda_i \quad , \quad U = \sum_{i=1}^{N} \lambda_i r_i \tag{1}$$

Since the aforementioned indices are unable to describe the reliability of a system properly, other reliability indices including System Average Interruption Frequency Index (SAIFI), System Average Interruption Duration Index (SAIDI), and Energy Not-Supplied (ENS) are represented as follow [21].

$$SAIFI = \frac{\sum_{i=1}^{N} \lambda_i N_i}{\sum_{i=1}^{N} N_i} \tag{2}$$

$$SAIDI = \frac{\sum_{i=1}^{N} U_i N_i}{\sum_{i=1}^{N} N_i} \tag{3}$$

$$ENS = \sum_{i=1}^{N} ENS_i \quad , \quad ENS_i = Lp_i f_i U_i \tag{4}$$

Where $N_i$ is the number of customers per load point, $ENS_i$ is the energy not-supplied at load point $i$, $Lp_i$ is peak load at load point $i$, and $f_i$ is load coefficient at load point $i$.

As mentioned earlier, the failure rate is not constant over time and at different environmental and operational conditions. However, in conventional reliability assessment, the aforementioned indices are calculated with a constant failure rate. In addition, assessing and improving the reliability of a distribution system based on the archived database is a reactive process which works based on past performance [18]. In other words, the system manager should wait for problems to occur then remedial actions for reliability improvement of the system are applied. Moreover, failure and repair time of each component of a distribution system are estimated using archived databases. However, in practice, such an archived database for all components are usually unavailable or inaccurate. Fuzzy sets theory is an efficient tool to handle such an uncertainty in reliability indices. Fuzzy logic models the environmental and operational conditions for each component in load point, mathematically and derives the aforementioned reliability indices. In this study, environmental and operational variables influencing the failure rate and repair time of components are considered as age of component, weather conditions, exposure to risk, and maintenance.

General structure of a fuzzy block includes three steps of fuzzifier, interface engine, and defuzzifier [22]. The fuzzifier receives information about effective environmental and operational variables for each component of distribution system and converts them to fuzzy values using membership functions. The membership functions generally include triangular or trapezoidal members [23]. A typical trapezoidal fuzzy number ($\tilde{A}$) is represented as $\tilde{A} = (a_1, a_2, a_3, a_4)$ which is calculated using fuzzy membership function of $\mu_{\tilde{A}}$ represented by (5). Note that the trapezoidal fuzzy number $\tilde{A}$ is a triangular number where $a_2$ and $a_3$ are the same in (5).

$$\mu_{\tilde{A}} = \begin{cases} 0 & x < a_1 \\ \dfrac{x - a_1}{a_2 - a_1} & a_1 < x \leq a_2 \\ 1 & a_2 < x \leq a_3 \\ \dfrac{a_3 - x}{a_3 - a_4} & a_3 < x \leq a_4 \\ 0 & x > a_4 \end{cases} \tag{5}$$

Where $a_1$ is the left foot, $a_2$ is the left peak, $a_3$ is the right peak, and $a_3$ the right foot of the fuzzy number. Depends on different conditions of an influential variable, several fuzzy numbers (triangular or trapezoidal) are defined in fuzzifier step.

The reverse process of fuzzifier is applied in the defuzzifier step which converts the output of inference engine into reliability index values. Here in this study, mean of maximum (mom) is applied in defuzzifier step. Inference engine is applied after fuzzifier step. In the inference engine, proper rules for reliability assessment are defined in the form of *if-then*, according to experts' experience and logic. As an example, *if* the exposure of line is *high* and the weather condition is *adverse*, *then* the failure rate is *high*. In addition, to derive reliability indices, the arithmetic operations on fuzzy numbers are required which are represented in (6) for two fuzzy numbers of $\tilde{A} = (a_1, a_2, a_3, a_4)$ and $\tilde{B} = (b_1, b_2, b_3, b_4)$. Note that the arithmetic operations in (6) can be applied for both trapezoidal and triangular numbers [24].

$$\tilde{A} \oplus \tilde{B} = (a_1 + b, a_2 + b_2, a_3 + b_3, a_4 + b_4)$$
$$\tilde{A} \ominus \tilde{B} = (a_1 - b_1, a_2 - b_2, a_3 - b_3, a_4 - b_4)$$
$$\tilde{A}/\tilde{B} = \left(\frac{a_1}{b_2}, \frac{a_2}{b_1}, \frac{a_3}{b_4}, \frac{a_4}{b_3}\right), a_i \text{ and } b_i > 0 \tag{6}$$
$$\tilde{A} \otimes \tilde{B} = (e_1, e_2, e_3, e_4)$$
$$\text{where } e_1 = \min T_1 = k^{th} \text{ element of } T_1$$

$$e_2 = max\ T_1 = l^{th}\ element\ of\ T_1$$
$$e_3 = k^{th}\ element\ of\ T_2$$
$$e_4 = l^{th}\ element\ of\ T_2$$
$$T_1 = \{a_1b_1, a_1b_2, a_2b_1, a_2b_2\}$$
$$T_2 = \{a_3b_3, a_3b_4, a_4b_3, a_4b_4\}$$

### III. FUZZY MODEL FOR COMPONENTS

In reliability analysis of distribution system, lines and transformers are considered the most critical components and the most influential environmental and operational variables on the failure rate of these components are the age of component, weather conditions, and the exposure to the risk. Accordingly, the following sections elaborate the fuzzy membership functions of theses influential variables, and failure rate and average outage (repair) time of each component.

#### A. Distribution Line

Distribution line, which is considered one of the most physically extensive yet susceptible components of the distribution system, is exposed to fail by several environmental or human-made reasons. In addition, these conditions may affect the outage time. Examples of these fault causing conditions are weather conditions (windy, snowy, stormy), mountainous area, forest area, urban area. Logically, when the weather condition is normal, the failure rate is less and repair time is shorter in compared with an adverse weather condition. Although for different operational cases, these variables may be considered different, *age*, *exposure*, and *weather conditions* are the most important variables influencing the failure rate and average outage time of distribution lines. Thus these variable are applied as fuzzy logic inputs.

##### 1) Age

Age is a key factor in the reliability assessment of a component. In a distribution system, the failure rate of distribution line rises as it gets older. Based on experts' experiences, a distribution is worn out when its age reached 30 years. Accordingly, as shown in Fig. 1, three fuzzy membership functions are defined as *young* which ranges between 0 to 8 years, *middle-aged* ranging between 4 to 22 years, and *old* which ranges between 18 to 30.

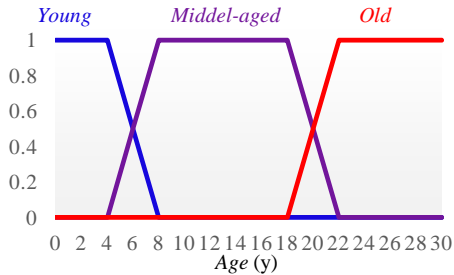

Figure 1. Fuzzy membership functions for line's age

##### 2) Exposure

The common samples of the line's exposure to the risk in distribution system which affect the reliability indices of failure rate and outage time are trees in the line path, mountainous area, and birds. Among them, exposure to the trees is the most common reason for failure in the distribution line. In this study, the exposure of the line to trees is estimated by percentage as shown in Fig. 2 where the exposure *low* is considered between 0 to 40 percent, while *average* and *high* are considered to intervals of 20 to 80 and 60 to 100 percent.

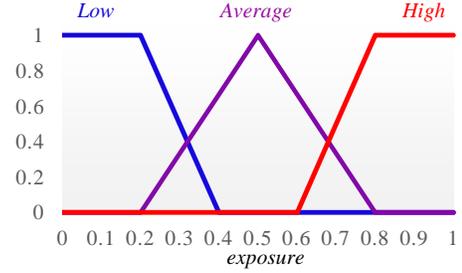

Figure 2. Fuzzy membership functions for line's exposure

##### 3) Weather Condition

The failure rate and outage time are strongly relevant to weather condition and they are increased in adverse weather condition. In this study, the weather condition as an input to the fuzzy model is classified into two categories of normal and adverse weather conditions [25]. Wind speed is a suitable representative to apply weather conditions in reliability assessment. Here in this study, as illustrated in Fig. 3, the wind speed of 0 to 80 km/h(≈50 mph) is considered *normal* while weather condition is considered *adverse* when the wind speed is at the intervals of 30 km/h(≈ **19** mph) to 100 km/h(≈ **62** mph).

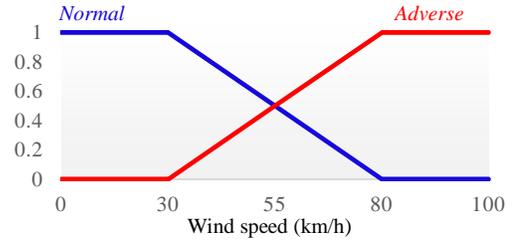

Figure 3. Fuzzy membership functions for weather condition

##### 4) Interruption Rate

The output variables of fuzzy logic for the reliability assessment of the distribution line is failure rate and average outage time. The fuzzy membership functions for the failure rate are depicted in Fig. 4. As shown, five membership functions are defined as *very low*, *low*, *average*, *high*, and *very high* to represent the failure rate, where *Very low* ranges from 0.8 to 1.2 while *low*, *average*, *high*, and *very high* are in 1.1 to .15, 1.4 to 2, 1.9 to 2.5, and 2.4 to 3, respectively.

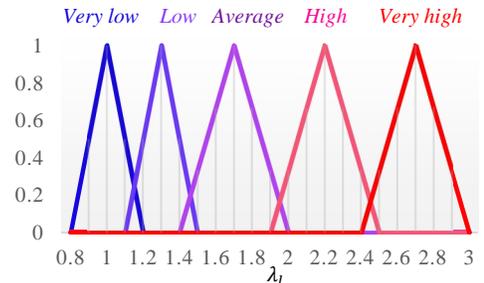

Figure 4. Fuzzy membership functions for failure rate of a distribution line

In addition, the fuzzy membership functions of average outage time (average repairing time) are represented by four functions of *good*, *suitable*, *bad*, and *very bad*. As shown Fig. 5, while the repairing time is considered *good* if it takes 4 to 6 hours, at the worst case i.e. *very bad*, it takes 12 to 20 hours to fix a failure. In other conditions, taking 5 to 9 hours for repairing a distribution line is considered as *suitable* and 7 to 15 hours is considered as a *bad* case.

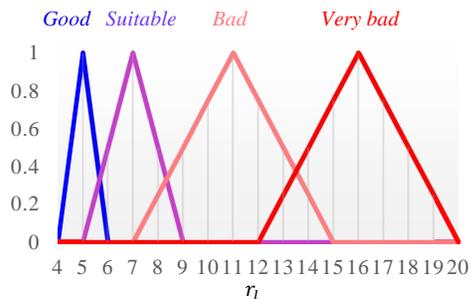

Figure 5. Fuzzy membership functions for average outage time of a distribution line

### B. Distribution Transformer

Transformer is one of the most valuable components of the distribution system whose failure may interrupt the supply for a large portion of demands. In addition to the high cost of transformers, the repairing time of transformers in compared with other components of the distribution system is significantly high. Accordingly, this component plays an important role in reliability assessment of a distribution system. In this study, the influential variables on the failure rate of a distribution transformer are considered *age*, *exposure*, and *operation condition*.

#### 1) Age

As mentioned earlier for the transmission line, the failure rate of a component rises when as it gets older [26]. Similar to line's age, three fuzzy membership functions of *young*, *middle-aged*, *old* are defined for the transformer's age where *young* ranges from 0 to 10 years, *middle-aged* ranges from 5 to 25 years, and a transformer with an age between 20 to 35 is considered as *old*.

#### 2) Exposure

Weather and geographical conditions of transformer' operation are also the other influential factor in the failure rate. Earthquake, lightning, flood, heavy rain, humidity, and salty dust are some examples of such conditions. Among them, moisture in oil is the most common reason of transformer's failure in short and log terms. Moisture causes breakdown in windings and it also increases discharges [27]. In this study, the amount of moisture in the oil of a transformer is considered as *exposure* to the risk index. The fuzzy membership functions for the *exposure* are three levels of *low* ranging from 0 to 40%, while *average* and *high* refer to the range of 20% to 80%, and 60% to 100%, respectively.

#### 3) Operation and Maintenance

In addition to the *age* and *exposure* variables, the *operation and maintenance* conditions of a transformer are influential factors on the failure rate. In other words, the failure rate in a *young* transformer which is operating at a *low* risk of moisture may still be high if it has been transferred, installed, or maintained inappropriately. Frequent overloading for long terms, lack of oil due to leaking, and improper oil are examples of improper operation and maintenance. In this study, the number of periodical check and test per year is considered as maintenance index and two fuzzy membership functions of *suitable* and *unsuitable* variables are defined which *suitable* means 3 to 10 times checking and testing per year while *unsuitable* refers to 0 to 7 times.

#### 4) Interruption Rate

The output variable of the reliability assessment of distribution transformer is failure rate represented by five membership functions as shown in Fig. 6. As shown, the failure rate of a transformer is *Very low* when it is between 0.8 to 1.2 whereas it is *Very high* between 1.9 to 2.5. The failure rate is *low, average*, or *high* when it is in ranges of 1.1 to 1.5, 1.3 to 1.7, or 1.6 to 2.2, respectively. Since the repairing time of a transformer is significantly higher than the repairing time of a line, this outage time is considered 200 hours on average.

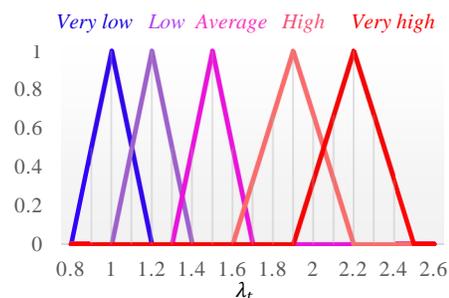

Figure 6. Fuzzy membership functions for transformer's failure rate

### IV. SIMULATION RESULTS

In this study, the aforementioned fuzzy membership functions of influential variables are applied in simulation on the IEEE RBTS Bus 2 network with MATLAB®. The IEEE RBTS Bus 2 network has 2 buses (33 kV and 11 kV), 22 transformers, and 4 feeders supplying 22 load points by 36 lines [19]. In this study, a complete series of possible combination cases of influential variables for the line and transformer are formed using the fuzzy membership functions and applied in inference engine. In inference engine, the reliability indices including SAIDI, SAIFI, and ENS are calculated using rule base and arithmetic operations on fuzzy numbers. In this paper, the result of reliability assessment for 9 samples of all possible cases is represented in Table 1. In this table, three numbers of the second column represent *age, exposure (trees in percent)*, and *weather condition* (wind (km/h)), respectively while three numbers of transformer represent *age, exposure(humidity)*, and *maintenance*. The fourth column of Table I represents improvement in variable(s) in compared with case 1 while case 1 is considered as a benchmark. As shown in Table II, the results of reliability assessment using fuzzy logic for the aforementioned cases are represented by five commonly used reliability indices represented in Section II. As seen, by implementing *young* components, decreasing the *exposure*'s risk of line and transformer, or by maintenance of the transformers, the reliability indices of the distribution system are enhanced. The illustration of this fact is shown in Fig. 7 for the SAIFI of the system.

TABLE I. INPUT VARIABLES OF FUZZY MEMBERSHIP FUNCTIONS

| Case | Line inputs | Transformer inputs | Improvement |
|---|---|---|---|
| 1 | (22, 0.85, 55) | (25, 0.5, 2) | benchmark |
| 2 | (22, 0.85, 25) | (25, 0.5, 2) | Weather condition |
| 3 | (22, 0.85, 55) | (25, 0.5, 5) | Transformer's maintenance |
| 4 | (22, 0.85, 55) | (15, 0.5, 2) | Transformer's age |
| 5 | (22, 0.65, 55) | (25, 0.5, 2) | Line's exposure |
| 6 | (10, 0.85, 55) | (25, 0.5, 2) | Line's age |
| 7 | (22, 0.65, 55) | (25, 0.5, 5) | Case 3 and 5 |
| 8 | (10, 0.85, 55) | (15, 0.5, 2) | Case 4 and 6 |
| 9 | (10, 0.15, 55) | (15, 0.5, 8) | Case 7 and 8 |

TABLE II. RELIABILITY ASSESSMENT USING FUZZY LOGIC

| Case | λ | U | SAIFI | SAIDI | ENS |
|---|---|---|---|---|---|
| 1 | 11.387 | 198.24 | 0.53548 | 8.5116 | 104770 |
| 2 | 8.9034 | 121.98 | 0.41897 | 5.7382 | 63063 |
| 3 | 8.872 | 115.68 | 0.4174 | 5.4238 | 59871 |
| 4 | 8.8285 | 106.98 | 0.41523 | 4.9893 | 55459 |
| 5 | 7.849 | 117.99 | 0.36951 | 5.5831 | 60890 |
| 6 | 6.9137 | 114.46 | 0.32565 | 5.4456 | 58962 |
| 7 | 6.8677 | 105.19 | 0.32335 | 4.983 | 54266 |
| 8 | 6.8388 | 99.459 | 0.3219 | 4.6967 | 51358 |
| 9 | 5.2734 | 78.832 | 0.24826 | 3.7289 | 40686 |

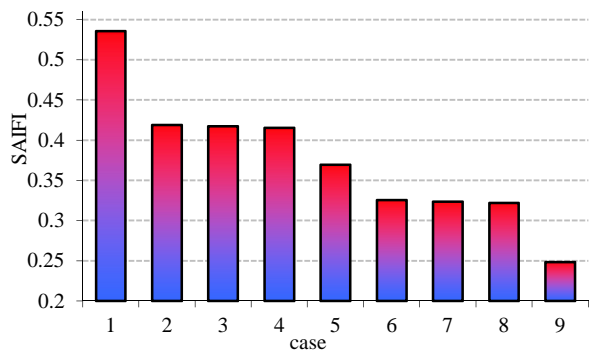

Figure 7. SAIFI assessment of IEEE RBTS Bus 2 system for different cases

In addition, as indicated in Table II and depicted in Fig. 8, among 5 cases where there is an improvement in only one of variables (case 2, 3, 4, 5, and 6), the average outage duration ($U$) has improved in case 4 more than other cases. In other words, in case 4, which represents using younger transformers (age of 15 years old) instead of older transformers (age of 22 years old), the $U$ is improved from 198.24 hours per year (case 1) to 106.98 hours per year (case 4).

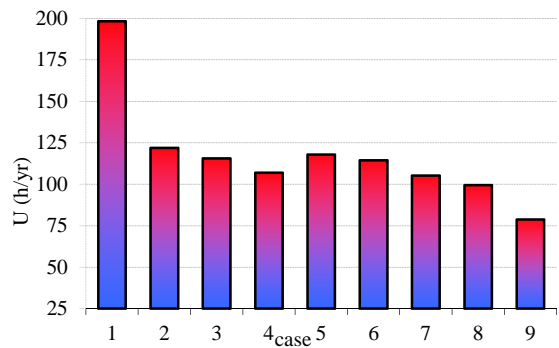

Figure 8. Average outage duration ($U$) index assessment of IEEE RBTS Bus 2 system for different cases using fuzzy logic

In addition, Figs. 9 and 10 give a better understanding of reliability improvement in the distribution system by enhancing the effect of the influential variables. These figures illustrate the percentage of improvement in case 2 to 9 in compared with case 1. As shown in Fig. 9, the ENS index of the system, which is a function of outage time and the amount of unsupplied energy, is improved from case 2 to case 9 where failure rate and outage time of components is decreased. Similar to the $U$, the ENS index is improved significantly (around 47%) for case 4 where the younger transformers are used instead of old ones.

Moreover, as inferred from Fig. 10, which illustrates the SAIFI improvement in other cases in compared with case 1, the reliability of the system can be improved in case 5 by 31% where *exposure* to the risk in the distribution line is decreased. In other words, only by cutting the tree limbs which have the risk of the conductors' contact, the reliability of the system can be improved, significantly.

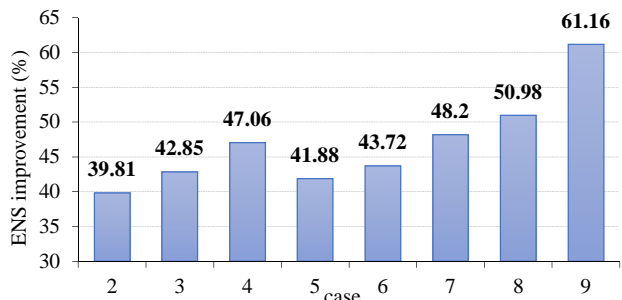

Figure 9. ENS improvement in other cases compared with case 1

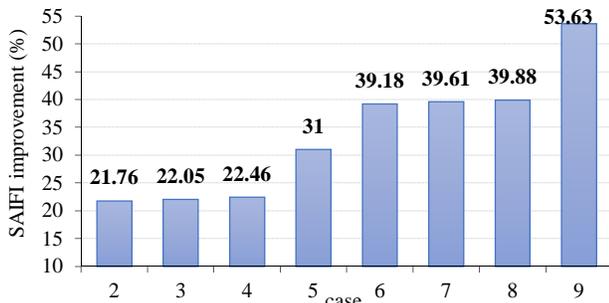

Figure 10. SAIFI improvement in other cases compared with case 1

## V. CONCLUSION

In this study, the influential variables on the reliability of the distribution system's components i.e. line and transformer are modeled by fuzzy logic theory. Such influential variables for a line are age, weather condition, and exposure to the risk while for a transformer are considered age, exposure, and maintenance. Using proper fuzzy membership functions and the rule base, the common reliability indices i.e. failure rate, average outage duration, SAIFI, SAIDI, and ENS are calculated by simulation on the IEEE RBTS Bus 2 system. In addition to the efficacy of the proposed fuzzy membership functions of influential variables, the results of simulation also indicate that only by using younger transformers, the reliability of the system can be improved, significantly. In addition, for this case study, by decreasing the exposure risk of lines in a distribution system, which is doable by cutting the trees limbs on the path of the lines, the reliability of the system is enhanced by 31%.